\documentclass[10pt,twocolumn,letterpaper]{article}

\usepackage{cvpr}
\usepackage{times}
\usepackage{graphicx}
\usepackage{amsmath}
\usepackage{amssymb}

\usepackage{authblk}

\usepackage[breaklinks=true,bookmarks=false]{hyperref}

\cvprfinalcopy 

\def\cvprPaperID{****} 

\begin{document}

\title{Streaming Networks: Enable A Robust Classification of Noise-Corrupted Images}

\author[1]{Sergey Tarasenko
}
\author[1]{Fumihiko Takahashi}
\affil[1]{
Next-Gen Mobility Division, Mobility R$\&$D Group, JapanTaxi
}

\maketitle

\begin{abstract}
The convolution neural nets (conv nets) have achieved a state-of-the-art performance in many applications of image and video processing. The most recent studies illustrate that the conv nets are fragile in terms of recognition accuracy to various image distortions such as noise, scaling, rotation, etc. In this study we focus on the problem of robust recognition accuracy of random noise distorted images. A common solution to this problem is either to add a lot of noisy images into a training dataset, which can be very costly, or use sophisticated loss function and denoising techniques. We introduce a novel conv net architecture with multiple streams. Each stream is taking a certain intensity slice of the original image as an input, and stream parameters are trained independently. We call this novel network a ``Streaming Net". Our results indicate that Streaming Net outperforms 1-stream conv net (employed as a single stream) and 1-stream wide conv net (employs the same number of filters as Streaming Net) in recognition accuracy of noise-corrupted images, while producing the same or higher recognition accuracy of no noise images in almost all of the tests. Thus, we introduce a new simple method to increase robustness of recognition of noisy images without using data generation or sophisticated training techniques.
\end{abstract}

\section{Introduction}

\subsection{Brief overview of the conv nets}
Since its first introduction in 1998 by Lecun et al. \cite{LeCun1998GradientbasedLA}, the convolutional neural networks (conv nets) have proved their effectiveness by achieving state-of-the-art solutions for many tasks.

There is a vast variety of conv net architectures can be found in the literature (AlexNet\cite{Krizhevsky2012ImageNetCW}, LeNet \cite{LeCun1998GradientbasedLA}, ResNet\cite{He2015DeepRL}, GoogLeNet \cite{Szegedy2014GoingDW}, VGG \cite{Simonyan2014VeryDC} etc). These are the networks that have a single stream structure, where information is processed consecutively layer-by-layer.

Recently, conv nets with more than one processing stream have started to gain popularity. To our knowledge, the first two-stream network was introduced by Chorpa \cite{Chopra2005LearningAS} and it is widely known as a ``Siamese network". The motivation behind two streams is that each of the streams carries information about a dedicated image. Images fed to the streams are different.

Most recently, two-stream networks have been used for the vast variety of recognition, segmentation and classification tasks such as

- similarity assessment (Siamese networks and pseudo-Siamese \cite{Chopra2005LearningAS, Zagoruyko2015LearningTC}); 

- change detection and classification \cite{Varghese2018ChangeNetAD}; 

- action recognition in videos \cite{Simonyan2014TwoStreamCN}; 

- one-shot image recognition \cite{Koch2015SiameseNN};

- simultaneous detection and segmentation \cite{Hariharan2014SimultaneousDA};

- human-object interaction recognition \cite{Gkioxari2017DetectingAR};

- group activity recognition \cite{Azar2018AMC}, etc.

\subsection{The conv nets and the primate brain}
The main peculiarity of the conv net is that in contrast to classical multilayer perceptron, conv net employs two types of layers in addition to non-linear activation ones. These are convolution and max-pooling layers. The convolution layers are layers of 2d filter kernels, which are tuned during the network training, while max-pooling layers are used for upsampling.

In the field of neuroscience, there two types of neurons in the primary visual cortex (V1) of the primate brain, which are proved to perform convolution and max-pooling operations. These are simple cells S and complex cells C, respectively. Originally, the common packaging of convolution layer followed by max-pooling layer, which is the gist of the conv nets, is the same as it is implemented in V1: within a single block simple cells S are followed by the complex cells C and the simple cells of the following block take inputs from the complex cells of the previous one \cite{Fukushima1980NeocognitronAS,Serre2005ATO, Serre2007RobustOR, Serre2005ObjectRW}.

One of the neuroscience-based architures was proposed by Serre et al. \cite{Serre2005ATO, Serre2007RobustOR, Serre2005ObjectRW}. The main difference between conv net and the network presented by Serre et al. is that in conv nets filters are tuned during network training, while in Serre et el. net's filter kernels are selected separately from classifier (SVM in this case) training. Therefore in this paper, we refer to the network proposed by Serre et al. as pseudo-conv net.
\begin{figure}[t]
\begin{center}
\includegraphics[width=1\linewidth]{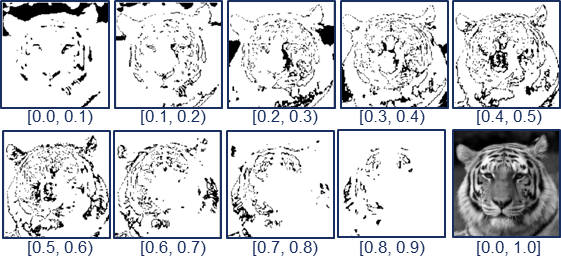}
\end{center}
\caption{Intensity Slices}
\label{fig:instesityslices}
\end{figure}


\begin{figure}[t]
\begin{center}
\includegraphics[width=1\linewidth]{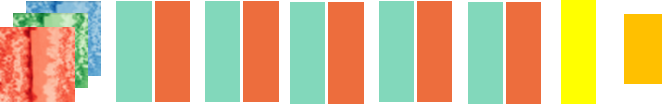}
\end{center}
\caption{One stream simple comv net}
\label{fig:stnet}
\end{figure}

Regarding the signal propagation in the brain networks, Thorpe at el. \cite{Thorpe1996SpeedOP,Thorpe2001SpikebasedSF,VanRullen2002SurfingAS} argued that the stronger the response of a given neuron, the faster such response should be produced, meaning that it takes less time to produce stronger output than a weaker one. 

Thorpe et al. have suggested that neural outputs produced nearly at the same time form waves of spikes. So even information about the static single image is propagated through the neural network in time separated packets called waves of spikes, thus a static image is unfolded in time due to different response time for stronger and weaker outputs.

\begin{figure}[t]
\begin{center}
\includegraphics[width=1\linewidth]{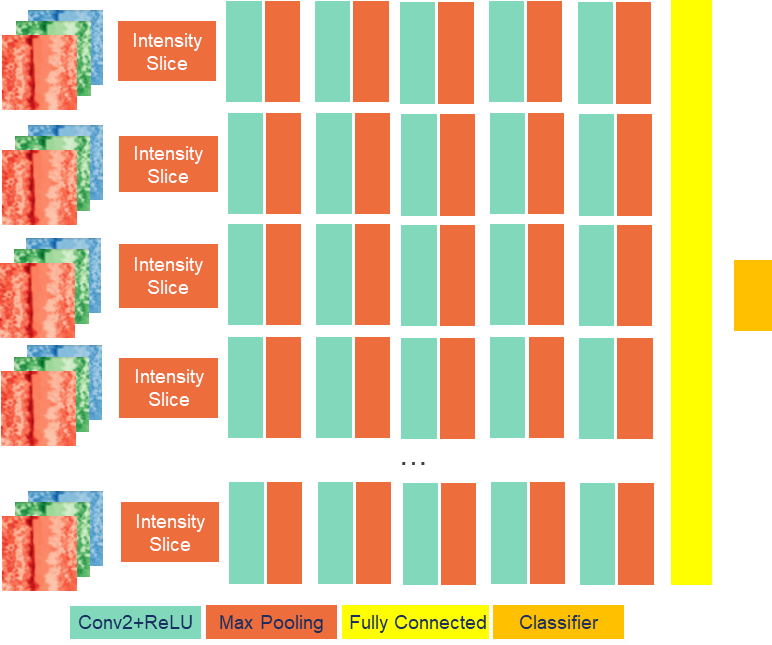}
\end{center}
\caption{Streaming Network architecture}
\label{fig:stnet}
\end{figure}

The idea of waves of spikes was then employed by Tarasenko \cite{TarasenkoWaves2011}. Tarasenko continued Work by Serre et al. by proposing an on-line learning method for feature extraction and extending the pseudo-conv net to implement a predictive coding \cite{Rao1999PredictiveCI} mechanism. 

The important peculiarity of work by Tarasenko is that to extract features images, containing complete information were used, while to enable mechanism of predictive coding after feature extraction, images were fed into the network by intensity slices (similar to waves of spikes). Examples of image intensity slices are presented in Fig. \ref{fig:instesityslices}. Every single image with normalized pixel values was split into 10 images, which correspond to one of the intensity slices ranging from 0.0 to 1.0 with step 0.1. Then these slices were consecutively propagated into the network.

\begin{figure}[t]
\begin{center}
\includegraphics[width=1.0\linewidth]{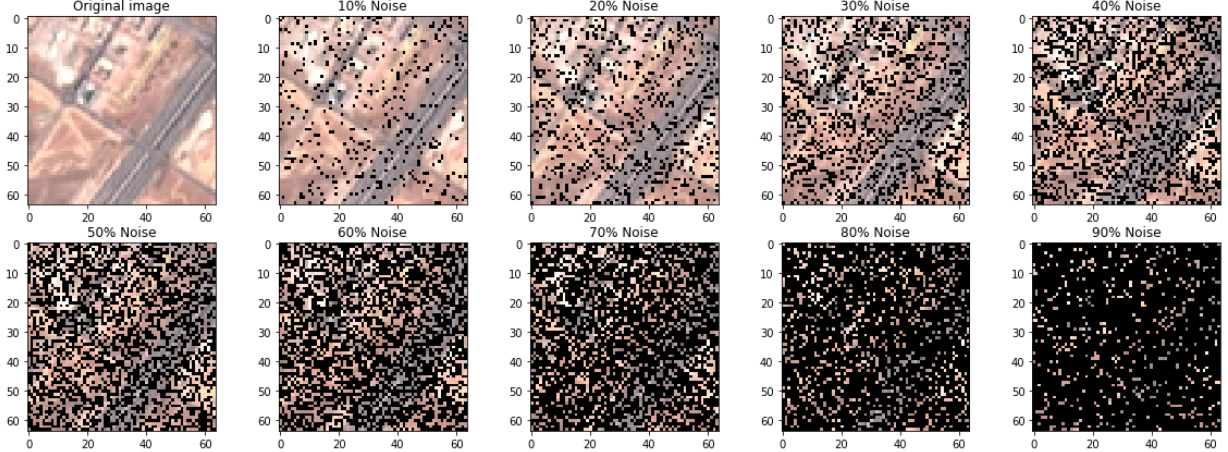}\end{center}
\caption{Adding random zero noise to Eurosat original data.}
\label{fig:noisy}
\end{figure}

\begin{figure*}[t]
\begin{center}
\includegraphics[width=1.0\linewidth]{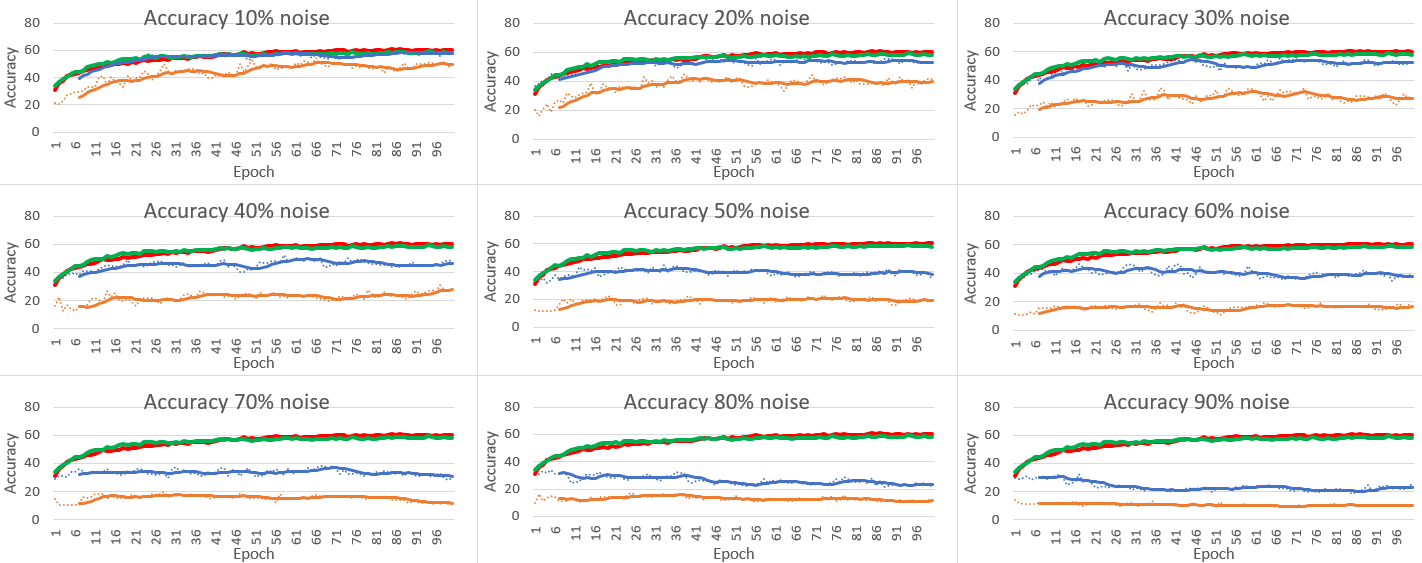}
\end{center}
\caption{Tests of Cifar10 dataset using Adam optimizer with 0.0005 learning rate for 1-stream conv net and 5-stream Streaming Net. Green line illustrates average prediction accuracy of no noise data by 5-stream Streaming Net, red line illustrates average prediction accuracy of no noise data by 1-stream network. Blue dotted line illustrates noise-corrupted data prediction accuracy (one sample run) by 5-stream Streaming Net and blue solid line is 7-point moving average smoothing. Orange dotted line illustrates noise-corrupted data prediction accuracy (one sample run) by 1-stream network and orange solid line is 7-point moving average smoothing.}
\label{fig:res_cifar10_0005}
\end{figure*}

\begin{figure*}[t]
\begin{center}
\includegraphics[width=1.0\linewidth]{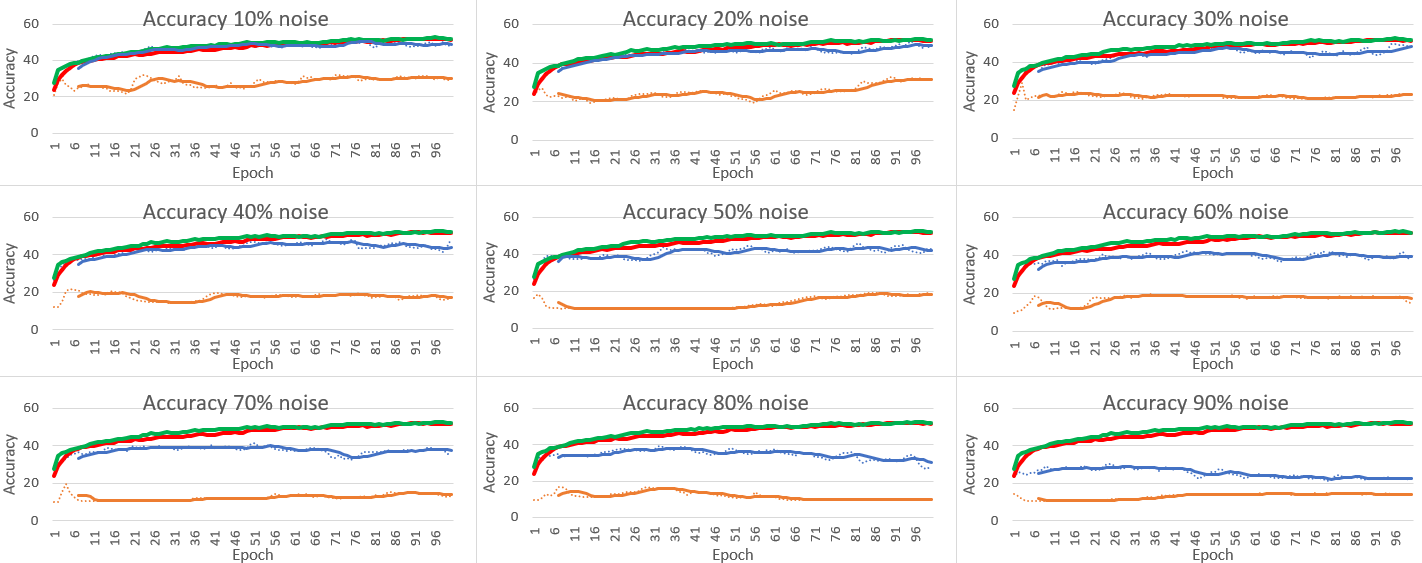}
\end{center}
\caption{Tests of Cifar10 dataset using Adam optimizer with 0.0001 learning rate for 1-stream conv net and 5-stream Streaming Net. The lines have the same meaning as in Fig. \ref{fig:res_cifar10_0005}}
\label{fig:res_cifar10_0001}
\end{figure*}

\begin{figure*}[t]
\begin{center}
\includegraphics[width=1.0\linewidth]{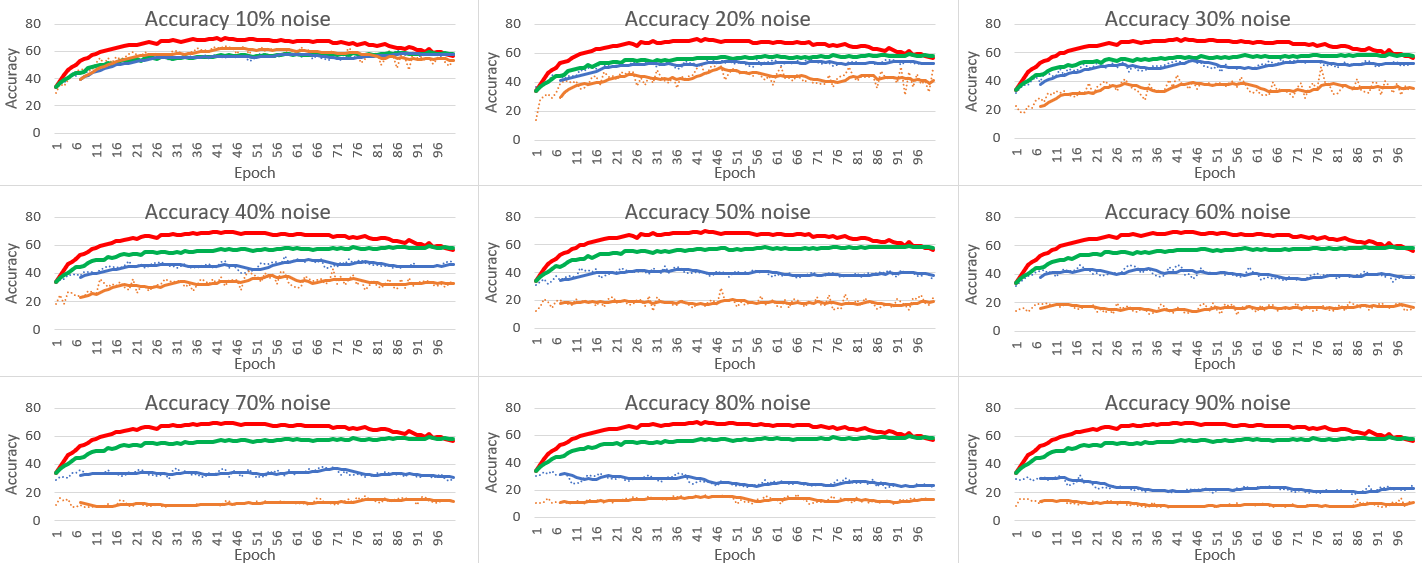}
\end{center}
\caption{Tests of Cifar10 dataset using Adam optimizer with 0.0005 learning rate for 1-stream wide conv net and 5-stream Streaming Net. The green, blue and blue dotted lines have the same meaning as in Fig. \ref{fig:res_cifar10_0005}, while other lines, corresponding to the same shape and color lines in Fig. \ref{fig:res_cifar10_0005}, refer to 1-stream wide conv net's performance.}
\label{fig:res_cifar10_fat_0005}
\end{figure*}

\begin{figure*}[t]
\begin{center}
\includegraphics[width=1.0\linewidth]{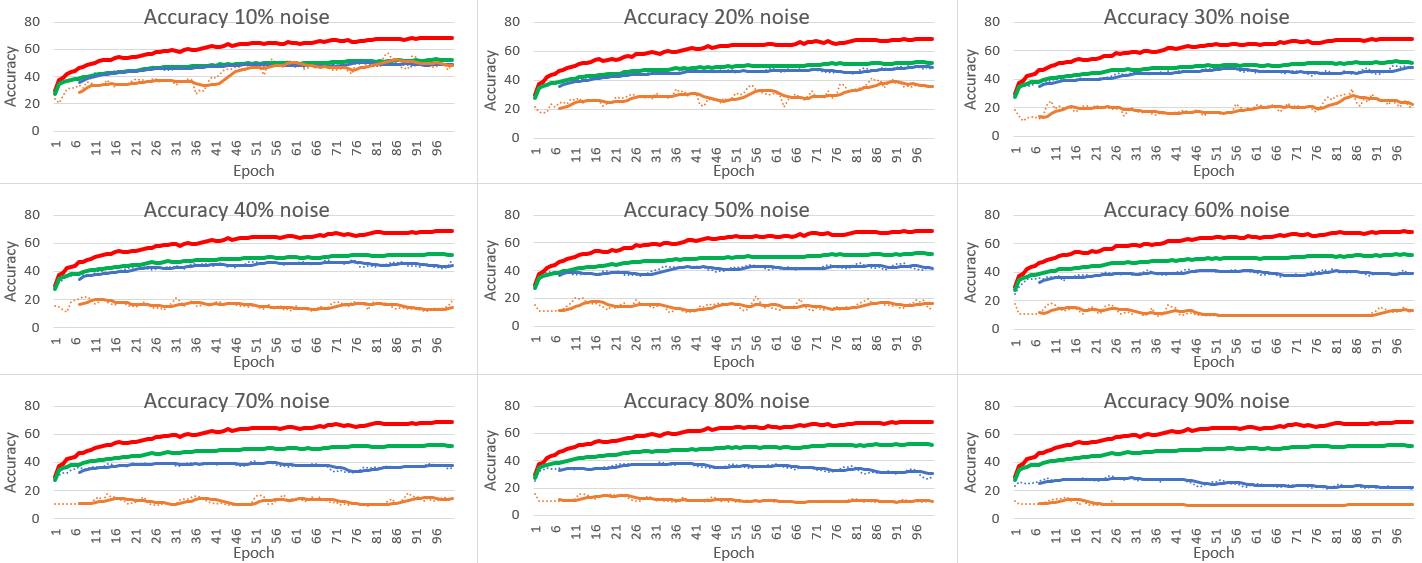}
\end{center}
\caption{Tests of Cifar10 dataset using Adam optimizer with 0.0005 learning rate for 1-stream wide conv net and 5-stream Streaming Net. The green, blue and blue dotted lines have the same meaning as in Fig. \ref{fig:res_cifar10_0005}, while other lines, corresponding to the same shape and color lines in Fig. \ref{fig:res_cifar10_0005}, refer to 1-stream wide conv net's performance.}
\label{fig:res_cifar10_fat_0001}
\end{figure*}

\begin{figure*}[t]
\begin{center}
\includegraphics[width=1.0\linewidth]{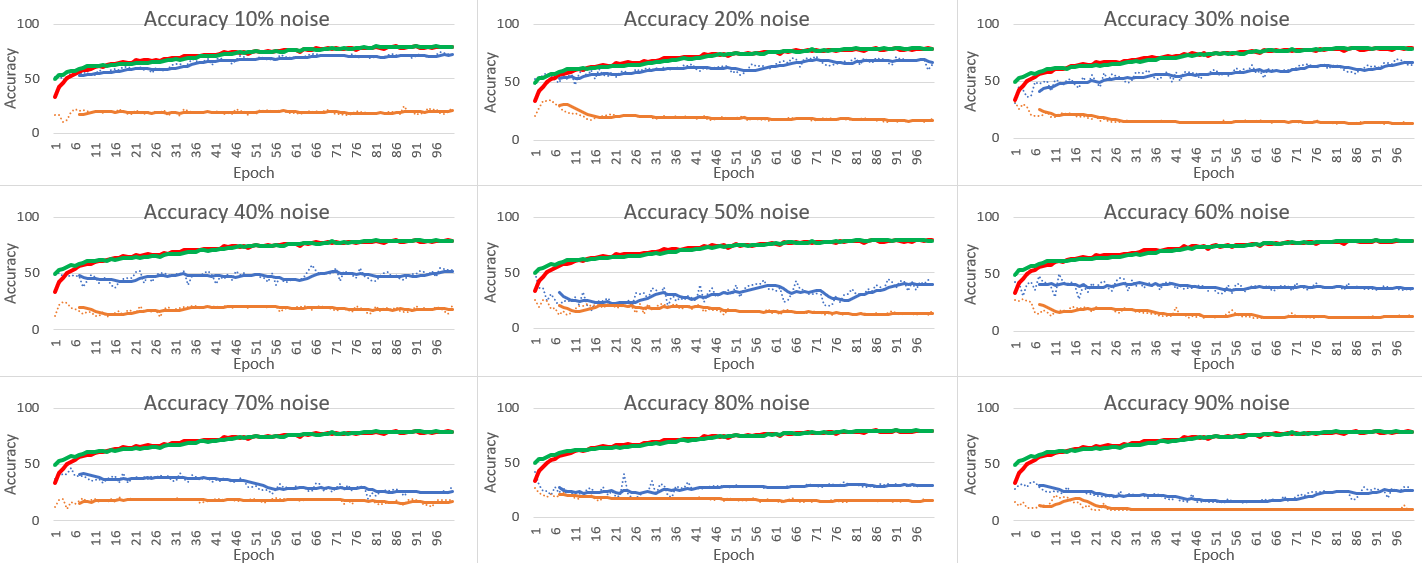}
\end{center}
\caption{Tests of Eurosat dataset using Adam optimizer with 0.0005 learning rate for 1-stream conv net and 5-stream Streaming Net. The lines have the same meaning as in Fig. \ref{fig:res_cifar10_0005}}
\label{fig:res_euro_0005}
\end{figure*}

\begin{figure*}[t]
\begin{center}
\includegraphics[width=1.0\linewidth]{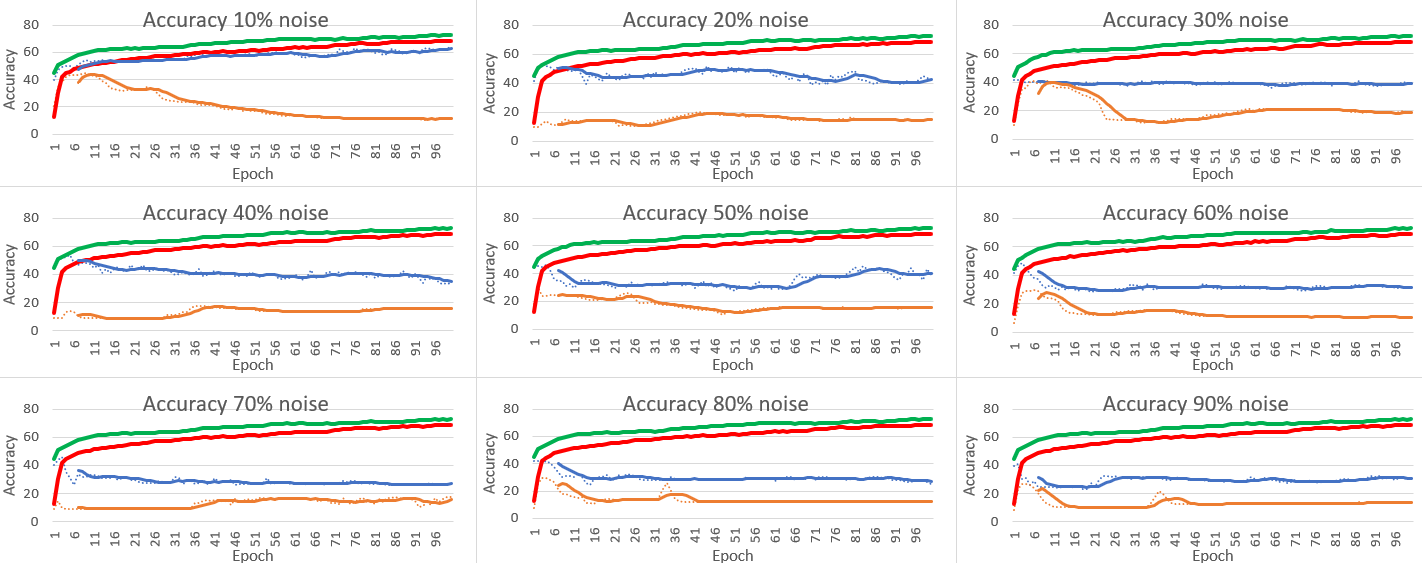}
\end{center}
\caption{Tests of Eurosat dataset using Adam optimizer with 0.0001 learning rate for 1-stream conv net and 5-stream Streaming Net. The lines have the same meaning as in Fig. \ref{fig:res_cifar10_0005}.}
\label{fig:res_euro_0001}
\end{figure*}

\begin{figure*}[t]
\begin{center}
\includegraphics[width=1.0\linewidth]{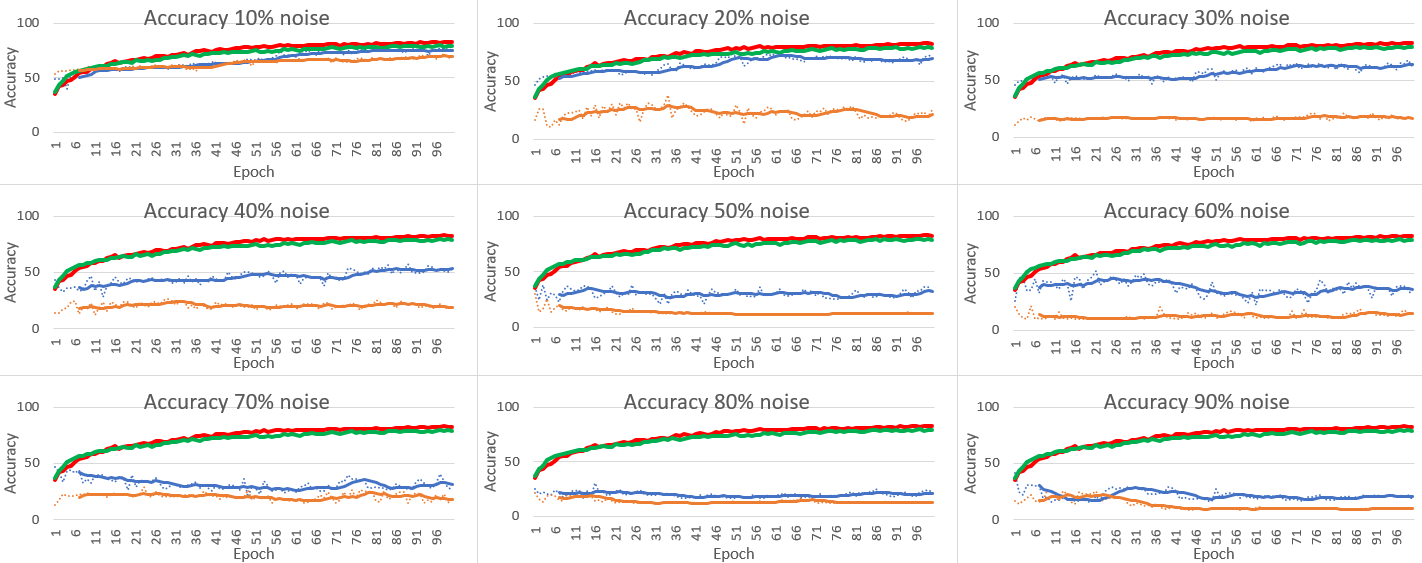}\end{center}
\caption{Tests of Eurosat dataset using Adam optimizer with 0.0005 learning rate for 1-stream wide conv net and 5-stream Streaming Net. The lines have the same meaning as in Fig. \ref{fig:res_cifar10_0005}, while other lines, corresponding to the same shape and color lines in Fig. \ref{fig:res_cifar10_0005}, refer to 1-stream wide conv net's performance.}
\label{fig:res_euro_fat_0005}
\end{figure*}

\begin{figure*}[t]
\begin{center}
\includegraphics[width=1.0\linewidth]{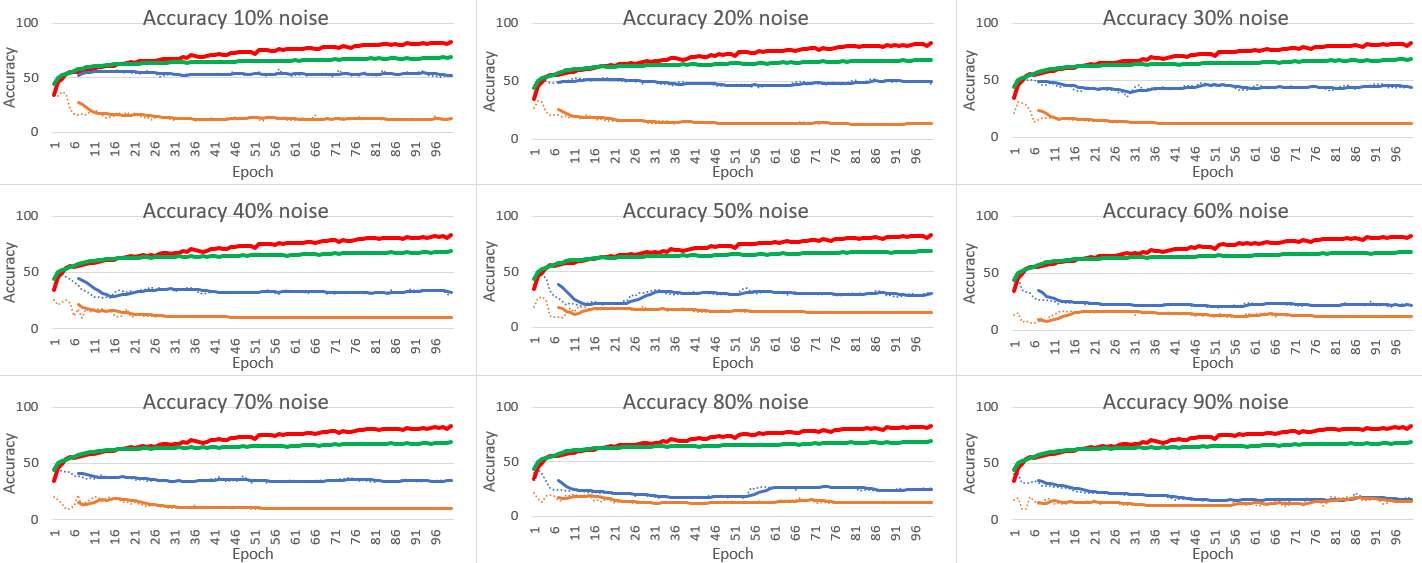}\end{center}
\caption{Tests of Eurosat dataset using Adam optimizer with 0.0001 learning rate for 1-stream wide conv net and 5-stream Streaming Net. The green, blue and blue dotted lines have the same meaning as in Fig. \ref{fig:res_cifar10_0005}, while other lines, corresponding to the same shape and color lines in Fig. \ref{fig:res_cifar10_0005}, refer to 1-stream wide conv net's performance.}
\label{fig:res_euro_fat_0001}
\end{figure*}

\begin{figure*}[t]
\begin{center}
\includegraphics[width=1.0\linewidth]{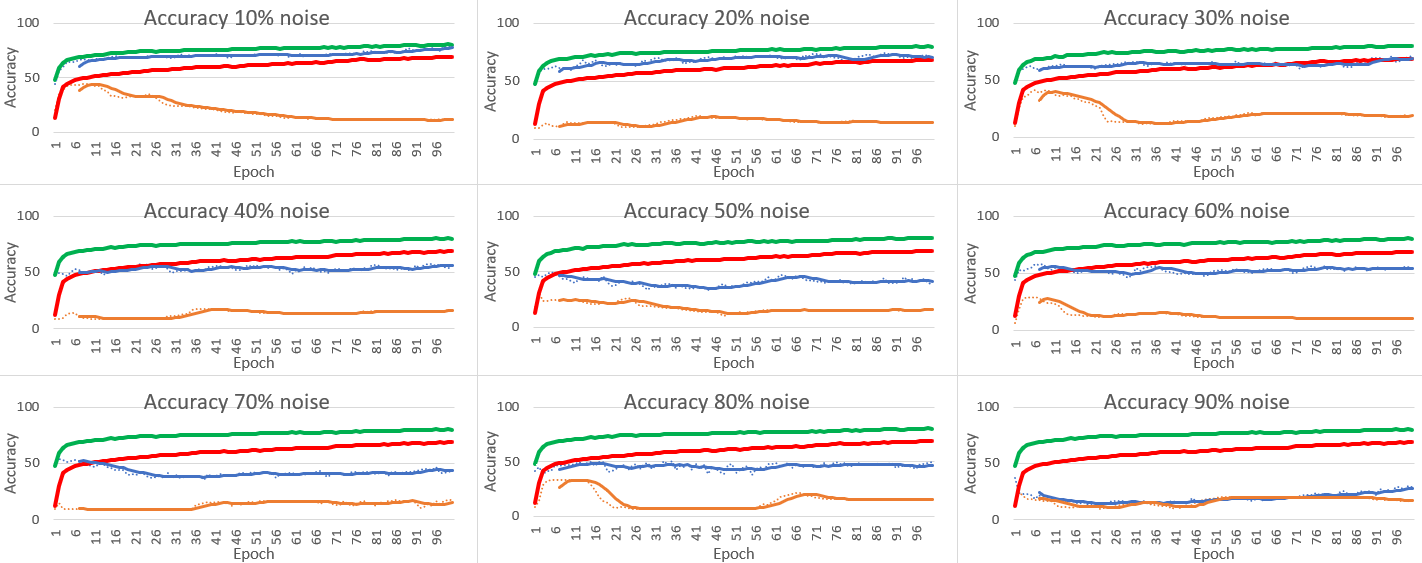}\end{center}
\caption{Tests of Eurosat dataset using Adam optimizer with 0.0001 learning rate for 1-stream wide conv net and 10-stream Streaming Net. The green, blue and blue dotted lines have the same meaning as in Fig. \ref{fig:res_cifar10_0005}, while other lines, corresponding to the same shape and color lines in Fig. \ref{fig:res_cifar10_0005}, refer to 1-stream wide conv net's performance.}
\label{fig:res_euro_10_0001}
\end{figure*}


\begin{figure*}[t]
\begin{center}
\includegraphics[width=1.0\linewidth]{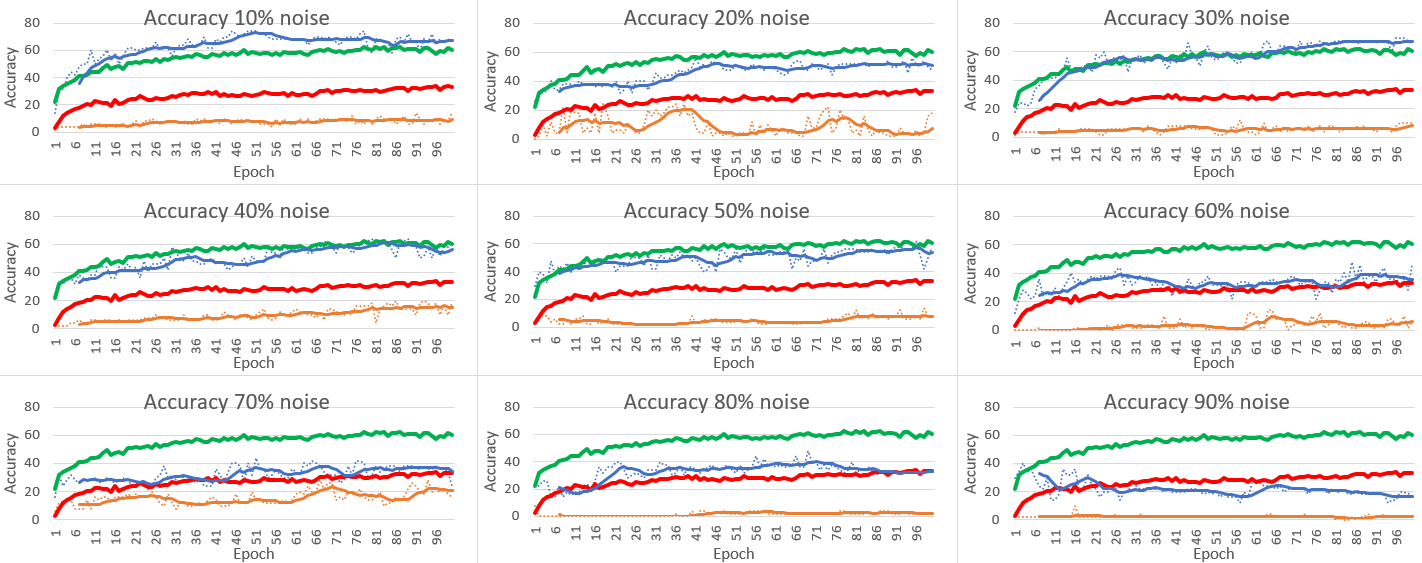}
\end{center}
\caption{Tests of UCmerced dataset using Adam optimizer with 0.0001 learning rate for 1-stream conv net and 5-stream Streaming Net. The lines have the same meaning as in Fig. \ref{fig:res_cifar10_0005}.}
\label{fig:res_ucmerced_0001}
\end{figure*}

\begin{figure*}[t]
\begin{center}
\includegraphics[width=1.0\linewidth]{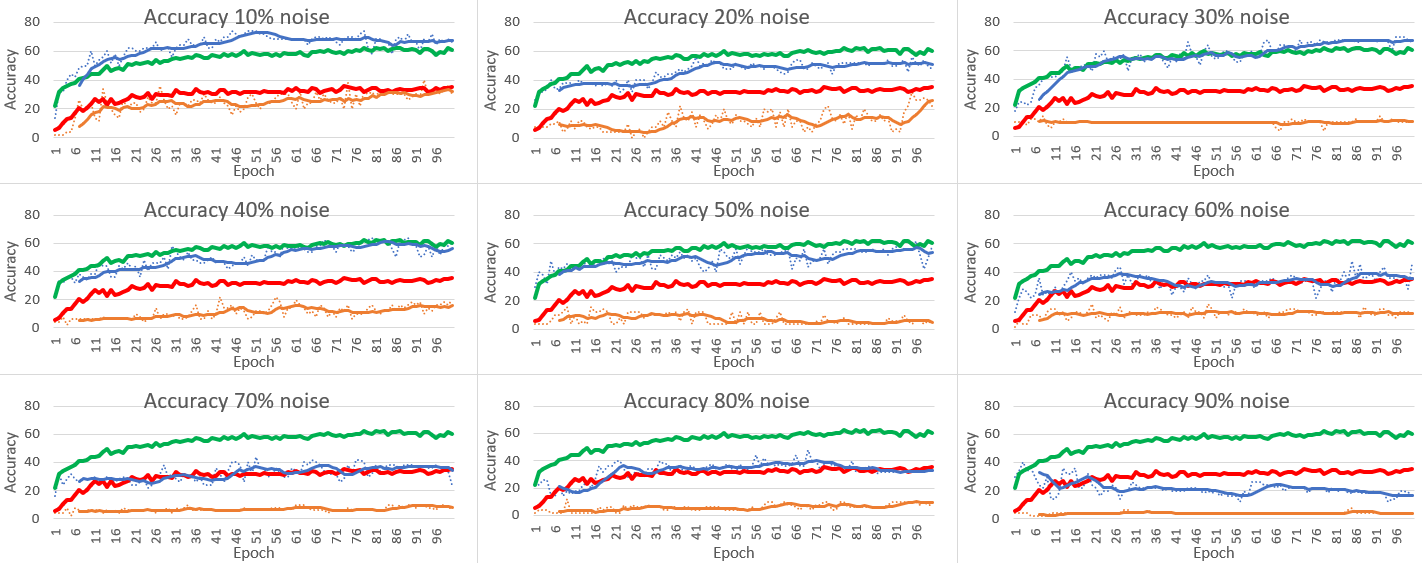}
\end{center}
\caption{Tests of UCmerced dataset using Adam optimizer with 0.0001 learning rate for 1-stream wide conv net and 5-stream Streaming Net. The lines have the same meaning as in Fig. \ref{fig:res_cifar10_0005}.}
\label{fig:res_ucmerced_fat_0001}
\end{figure*}

\subsection{The conv nets and image distortions}
Although conv nets have conquered the glory of state-of-the-art solutions in image and video processing, recent studies have illustrated that conv nets' performance is extremely fragile for distortions of the input images such as noise, image occlusions, rotation, scaling, etc. \cite{Szegedy2013IntriguingPO, Eykholt2017RobustPA}. In this paper, we will approach the issue of robust recognition when images are corrupted with random zero-noise when a certain portion of pixels across the entire image is randomly set to zero intensity value.

\section{Related Work}
The topic of robust recognition by conv nets under conditions of noise has been explored in \cite{Aghdam2016AnalyzingTS,Aghdam2016IncreasingTS}.

In work \cite{Aghdam2016AnalyzingTS}, authors analyzed the robustness (stability) of conv nets against image degradation due to noise. The same group of authors suggests a method to increase the stability of conv nets by introducing a denoising layers \cite{Aghdam2016IncreasingTS}.

In this paper, we concentrate on the issue of robust recognition under conditions that images are corrupted by random zero-noise, i.e., random image pixels are set to zero.

In this paper, we concentrate on multi-stream networks having their inputs is different intensity slices of the same image. In particular, we test the five-stream network with each stream having intensity slices as an input. The image is cut into intensity slice neither to have overlapping slices nor to miss any of intensity scales.


\section{Network Architecture: Streaming Net}

Here we introduce a novel conv net architecture.

We take a 1-stream conv net and add image intensity slicing module as an input layer of the network. Then we clone new networks with intensity slices set to extract different intensity slices. Each such network constitutes a single stream. Finally, we concatenate outputs of all the streams to one fully connected layer, which is connected with a classifier. A number of fully connected layers after concatenated layer can vary. 

The weights and biases within every single stream are not coupled with weights and biases of the other streams and trained independently.

We call this novel architecture a Streaming Net. The architecture of the Streaming Net is presented in Fig. \ref{fig:stnet}. The reason to select the name ``Streaming Net" is that besides for the obvious fact that such a network has multiple parallel streams, we also choose the word ``streaming" by analogy with parallel streams, when running Nvidia GPUs \cite{Sanders2010CudaBE} or TensorFlow\footnote{https://www.tensorflow.org/} distributed processing streams \cite{Abadi2016TensorFlowAS}. Each stream processes a designated piece of data asynchronously, thus enabling parallel processing and computation essential speed up.


\section{Experiments}

For our experiments we use three data sets. The selected datasets are cifar10\footnote{https://www.cs.toronto.edu/~kriz/cifar.html}, Eurosat (rgb)\footnote{https://github.com/phelber/eurosat} \cite{Helber2017EuroSAT} and UCmerced land use\footnote{http://weegee.vision.ucmerced.edu/datasets/landuse.html}.

For all our experiments, we use Adam optimizer with learning rate of values 0.0005 and 0.0001 accompanied by $\beta$1 = 0.99, $\beta$2 = 0.9 and $\epsilon$ = 1e-08, and run all the trainings for 100 epochs. For the UCmerced land-use dataset we used only 0.0001 learning rate as only this value enable network to converge to high level of accuracy.

For each dataset we run the networks for noise level (ratio of pixels corrupted with noise) ranging from 0.1 to 0.9 with step 0.1, thus constituting 9 different levels. Examples of different noise levels for a selected Eurosat image are illustrated in Fig.\ref{fig:noisy}.

Throughout the experiments, we use SoftMax classifier. 

When we train the network after each iteration we compute network accuracy for test data without noise and the test data corrupted with noise.

We consider the performance of three networks:

1) A 1-stream simple conv net;

2) Streaming Net with 5 streams with intensity slices [0.0,0.2), [0.2,0.4), [0.4,0.6), [0.6,0.8) and [0.8,1.1);

3) A 1-stream wide conv net, which is obtained from 1-stream simple conv net by multiplying number of filters in all convolution layers by a factor of 5 (the same as the number of streams in the Streaming Net).

A 1-stream simple conv net has the following structures: 1) conv layer with 32 7x7 filters plus ReLU activation and 2x2 Max-pooling layers;
2) conv layer with 64 5x5 filters plus ReLU activation and 2x2 Max-pooling layers;
3) conv layer with 128 3x3 filters plus ReLU activation and 2x2 Max-pooling layers;
4) conv layer with 256 1x1 filters plus ReLU activation and 2x2 Max-pooling layers;
5) conv layer with 4\footnote{For cifar10 dataset we use 10 1x1 filters} 1x1 filters plus ReLU activation and 2x2 Max-pooling layers;
6) fully connected layer;
7) SoftMax layer with the number of output neurons corresponding to the number of classes.

To train the networks, we use AWS p3.x2large\footnote{https://aws.amazon.com/ec2/instance-types/p3/} instances with Nvidia V100 GPU on-board and Nvidia GPU Cloud (NGC)\footnote{https://www.nvidia.com/en-us/gpu-cloud/} AIM with TensorFlow deployed with Docker container\footnote{https://ngc.nvidia.com/catalog/containers/nvidia:tensorflow}.

\subsection{Cifar10 tests}

Cifar10 dataset contains RGB 32x32 images of 10 classes (airplane, automobile, bird, cat, deer, dog, frog, horse, ship, truck). The total number of images is 60,000 with 6,000 images for each class. To train and test the networks, we use 50,000 and 10,000 images respectively.

For this dataset we observe, that Streaming Net exhibits the same accuracy for images without any noise, while completely dominates over 1-stream simple conv net across all the noise levels (Figs. \ref{fig:res_cifar10_0005} and \ref{fig:res_cifar10_0001}).

However 1-stream wide conv net beats Streaming Net in term of no noise image recognition, Streaming Net is dominating in recognition rate of noise images everywhere. Only in the case of 10$\%$ noise both networks exhibit comparable recognition accuracy of noise images (Figs. \ref{fig:res_cifar10_fat_0005} and \ref{fig:res_cifar10_fat_0001}).

\subsection{Eurosat RGB tests}

Eurosat dataset contains Sentinel-2 satellite images covering both 13 spectral bands and RGB (3-channel) 64x64 images and consisting of 10 classes (AnnualCrop, Forest, HerbaceousVegetation, Highway, Industrial, Pasture, PermanentCrop, Residential, River, SeaLake) with in total 27,000 labeled and geo-referenced images. We use only RGB images for this study.

For this dataset we observe that Streaming Net dominates over 1-stream conv net for both no noise images and across all the noise levels (Figs. \ref{fig:res_euro_0005} and \ref{fig:res_euro_0001}).

A 1-stream wide conv net beats Streaming Net in terms of no noise image recognition accuracy only for a small learning rate of 0.0001, while for learning rate 0.0005 the performance of both networks is nearly the same. On the other hand, Streaming Net is dominating in recognition rate of noisy images for all the noise levels (Figs. \ref{fig:res_euro_fat_0005} and \ref{fig:res_euro_fat_0001}).

\subsection{UCmerced Land Use tests}

The UCmerced land-use dataset contains 256x256 RGB satellite images of 21 class of land use. There are 100 images for each of the following classes: agricultural, airplane, baseballdiamond, beach, buildings, chaparral, denseresidential, forest, freeway, golfcourse, harbor, intersection, mediumresidential, mobilehomepark, overpass, parkinglot, river, runway, sparseresidential, storagetanks, tenniscourt.

This dataset contains images of the biggest size used in our test. For UCmerced land use data 5-stream Streaming Net dominates over both 1-stream simple conv net (Fig.\ref{fig:res_ucmerced_0001}) and 1-stream wide conv net (Fig.\ref{fig:res_ucmerced_fat_0001}), leaving them far behind in terms of recognition accuracy.

It is also important to emphasize the recognition accuracy curve for Streaming Net stays above the recognition accuracy of no noise images for 1-stream simple conv net up to the 80$\%$ of noise (Fig.\ref{fig:res_ucmerced_0001}).

\subsection{Streaming Net with more streams}
\label{many_streams}

\textbf{Adding a stream with no-noise image.} Throughout the datasets, we have also tested Streaming Net with 6 streams, when the sixth stream had the whole image as the input. Our results indicated not significate performance gain for either of the datasets.

\textbf{Use more precise intensity slices.} For Eurosat dataset, we have also trained 10-stream Streaming Net with a learning rate 0.0001 and intensity slices [0.0,0.1), [0.1,0.2), [0.2,0.3), [0.3,0.4), [0.4,0.5), [0.5,0.6), [0.6,0.7), [0.7,0.8), [0.8,0.9) and [0.9,1.1). The results are presented in Fig.\ref{fig:res_euro_10_0001}. 

\begin{figure}[t]
\begin{center}
\includegraphics[width=1.0\linewidth]{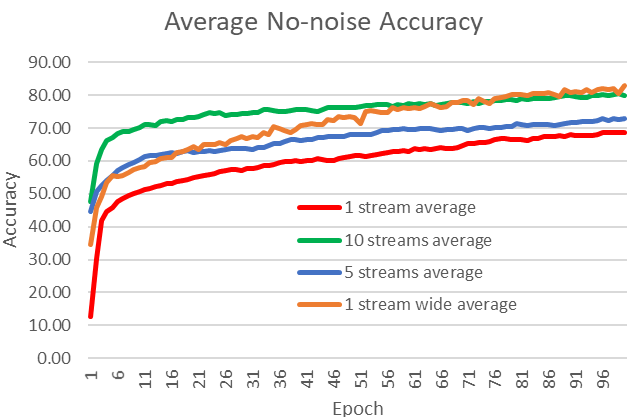}
\end{center}
\caption{Average recognition accuracy curves (learning curves) for the Eurosat dataset with a learing rate 0.0001.}
\label{fig:euso_avr}
\end{figure}

The results imply for no-noise image recognition accuracy indicate the following conclusions (Fig. \ref{fig:euso_avr}): 

1) recognition accuracy for the 10-stream Streaming Net using no-noise data is higher than the one for 5-stream Streaming Net and 1-stream simple conve net; 

2) recognition accuracy for the 10-stream Streaming Net and 1-stream wide network converges to the save level, however 10-stream Streaming Net convergence faster;

3) the 10-stream Streaming Net converges faster than other networks;

4) the 1-stream wide conv net converges faster than 5-stream Streaming Net converges and 1-stream simple conv net; 

5) the 5-stream Streaming Net converges faster than 1-stream simple conv net.

Finally, one can infer from Figs. \ref{fig:res_euro_0001}, \ref{fig:res_euro_fat_0001} and \ref{fig:res_euro_10_0001} that the 10-stream Streaming Net has the highest recognition accuracy across all the noise levels.

\section{Discussion and Conclusion}

In this paper, we have tested performance of 1-stream simple conv net, 1-stream wide conv net, and a newly introduces Streaming Net.
The results of performance comparison between 1-stream simple conv net and Streaming Net are as follows:

1) adding streams containing the same network with the input of different image transformations provides the same or better performance for uncorrupted data, while exhibiting extremely robust recognition under various noise levels;

2) for complex datasets containing large images, Streaming Net is the only one capable of high recognition accuracy for both noise-corrupted and uncorrupted images;

3) adding a stream which takes the whole image as an input, has no significant effect;

4) adding more stream with more precise slicing boost recognition accuracy for both no-noise and noise-corrupted images.

As we have emphasized in the section \ref{many_streams}, the recognition accuracy increasing from 1-stream simple conv net to 5-stream Streaming Net and to the 1-stream wide conv net and 10-stream Streaming Net (Streaming Net converges faster).
We suggest that the boost (higher accuracy and faster convergence, Fig. \ref{fig:euso_avr}) in recognition accuracy is caused by increasing diversity of filters as sugested by Lecun et al. \cite{LeCuneffprop98}. 

The introduction of the 1-stream wide network illustrates that for moderately small image size (cifar10) increasing number of filters in convolution layers can indeed boost no noise image cecognition accuracy compared to the Streaming Net. For the Eurosat dataset, this, however, is only true for the case of a small learning rate (0.0001), while for a learning rate of 0.0005 both 1-stream wide conv net and Streaming Net show the same performance. However, in the case of large image size (UCmerced) 1-stream wide conv net and 1-stream conv net are both far behind the Streaming Net.

Regarding the noise levels, the Streaming Net is dominating both 1-stream and 1-stream wide conv net, the only exception is for 10$\%$ noise case for cifar10 dataset, where both wide conv net and Streaming Net are exhibiting comparable performance.

To date, the common solution for dealing with noise in the images was to generate many noisy images during training a conv net. However, this is a very costly approach. In study \cite{Aghdam2016IncreasingTS} , it was suggested to use a new objective function regularized by local Lipschitz constant and to train ReLU layer for restoring noise images.

In this study, we illustrate that it is possible to achieve noise-resistant performance by using a multi-stream network with inputs of different intensity slices without any noise or using additional denoising techniques or complex objective functions.

For future studies, we aim to employ state-of-the-art architecture like AlexNet, VGG-family and ResNet-family networks as a single stream in the Streaming Net. 

Furthermore, it may be reasonable to use Streaming Nets as a single stream for the bigger Streaming Nets in some cases. 

Finally, thoughout this study we have used non-intersecting intensity slices of images, it could be an interesting challenge to use both non-intersecting and intersecting image slices and results of the various transformation of such slices as inputs to the Streaming Net.

To summarize, we emphasize that the main achievement of our work is the introduction of a simple and uncostly method to increase conv net robustness against the noise without using complex data generation techniques or sophisticated learning algorithms.




{\small

\begin{thebibliography}{9}
\bibitem{Abadi2016TensorFlowAS}
M. Abadi, P. Barham, J. Chen, Z. Chen, A. Davis, J. Dean, M. Devin, S. Ghemawat, G. Irving, M. Isard, M. Kudlur, J. Levenberg, R. Monga, S. Moore, D. G. Murray, B. Steiner, P. A. Tucker, V. Vasudevan, P. Warden, M. Wicke, Y. Yu and X. Zheng
(2016) TensorFlow: A System for Large-Scale Machine Learning, \textit{OSDI 2016}.

\bibitem{Azar2018AMC}
S. M. Azar, M. G. Atigh and A. Nickabadi (2018) A Multi-Stream Convolutional Neural Network Framework for Group Activity Recognition, \textit{ArXiv, abs/1812.10328}

\bibitem{Aghdam2016AnalyzingTS}
H. H. Aghdam and E. J. Heravi and D. Puig (2016) Analyzing the Stability of Convolutional Neural Networks against Image Degradation, \textit{VISIGRAPP}.

\bibitem{Aghdam2016IncreasingTS}
H. H. Aghdam, E. J. Heravi and D. Puig (2016) Increasing the Stability of CNNs using a Denoising Layer Regularized by Local Lipschitz Constant in Road Understanding Problems, \textit{VISIGRAPP}.

\bibitem{Chopra2005LearningAS}
S. Chopra and R. Hadsell and Y. LeCun (2005) Learning a similarity metric discriminatively, with application to face verification, \textit{IEEE Computer Society Conference on Computer Vision and Pattern Recognition (CVPR 05)}, p. 539-546.

\bibitem{Eykholt2017RobustPA}
K. Eykholt, I. Evtimov, E. Fernandes, B. Li, A. Rahmati, C. Xiao, A. Prakash, T. Kohno and D. Song (2018) Robust Physical-World Attacks on Deep Learning Models, \textit{CVPR 2018}.
\bibitem{Fukushima1980NeocognitronAS}
K. Fukushima (1980) Neocognitron: A self-organizing neural network model for a mechanism of pattern recognition unaffected by shift in position,
\textit{Biological Cybernetics}, 36, p. 193-202.

\bibitem{Gkioxari2017DetectingAR}
G. Gkioxari, R. B. Girshick, P. Doll{\'a}r and K. He (2018) Detecting and Recognizing Human-Object Interactions, \textit{IEEE/CVF Conference on Computer Vision and Pattern Recognition}, p. 8359-8367.

\bibitem{Hariharan2014SimultaneousDA}
B. Hariharan and P. Andr{\'e}s Arbel{\'a}ez and R. B. Girshick and J. Malik (2014) Simultaneous Detection and Segmentation, \textit{ECCV 2014}

\bibitem{He2015DeepRL}
K. He, X. Zhang, S. Ren and J. Sun (2015) Deep Residual Learning for Image Recognition, \textit{IEEE Conference on Computer Vision and Pattern Recognition (CVPR 16)}, p. 770-778.

\bibitem{Helber2017EuroSAT}
P. Helber, B. Bischke, A. Dengel and D. Borth (2017) EuroSAT: A Novel Dataset and Deep Learning Benchmark for Land Use and Land Cover Classification, \textit{IEEE Journal of Selected Topics in Applied Earth Observations and Remote Sensing}, 12, p. 2217-2226.

\bibitem{Koch2015SiameseNN}
G. R. Koch (2015) Siamese Neural Networks for One-Shot Image Recognition

\bibitem{Krizhevsky2012ImageNetCW}
A. Krizhevsky, I. Sutskever and G. E. Hinton (2012) ImageNet Classification with Deep Convolutional Neural Networks, \textit{NIPS 2012}.

\bibitem{LeCun1998GradientbasedLA}
Y. LeCun, L. Bottou, Y. Bengio and P. Haffner (1998) Gradient-based learning applied to document recognition. \textit{Proceeding of the IEEE}, 86 (11), p. 2278-2324.

\bibitem{LeCuneffprop98} 
Y. A. Lecun, L. Bottou, G. B. Orr and K.-R. M{\"u}ller (1998) Efficient backprop, \textit{In Eds. Orr, G. and M{\"u}ller K. Neural Networks: Tricks of the trade}, Springer, p. 9-48.

\bibitem{Rao1999PredictiveCI}
R. P. N. Rao and D. H. Ballard (1999) {Predictive coding in the visual cortex: a functional interpretation of some extra-classical receptive-field effects}, \textit{Nature Neuroscience}, 2, p. 79-87.

\bibitem{Sanders2010CudaBE}
J. Sanders and E. Kandrot (2010) Cuda by Example: An Introduction to General-Purpose Gpu Programming.

\bibitem{Serre2007RobustOR}
T. Serre, L. Wolf, S. M. Bileschi, M. Riesenhuber and T. A. Poggio (2007) Robust Object Recognition with Cortex-Like Mechanisms, \textit{IEEE Transactions on Pattern Analysis and Machine Intelligence}, 29, p. 411-426.

\bibitem{Serre2005ObjectRW}
T. Serre, L. Wolf and T. A. Poggio (2005) Object recognition with features inspired by visual cortex, \textit{IEEE Computer Society Conference on Computer Vision and Pattern Recognition (CVPR 05)}, p. 994-1000.

\bibitem{Serre2005ATO}
T. Serre, M. Kouh, C. F. Cadieu, U. Knoblich, G. Kreiman and T. Poggio (2005) A Theory of Object Recognition: Computations and Circuits in the Feedforward Path of the Ventral Stream in Primate Visual Cortex.

\bibitem{Simonyan2014VeryDC}
K. Simonyan and A. Zisserman (2014) Very Deep Convolutional Networks for Large-Scale Image Recognition, Arxiv, abs/1409.1556.

\bibitem{Simonyan2014TwoStreamCN}
K. Simonyan and A. Zisserman (2014) Two-Stream Convolutional Networks for Action Recognition in Videos,
\textit{ArXiv, abs/1406.2199}.

\bibitem{Szegedy2013IntriguingPO}
C. Szegedy and W. Zaremba and I. Sutskever and J. Bruna and D. Erhan and I. J. Goodfellow and R. Fergus (2013) Intriguing properties of neural networks, \textit{Arxiv, abs/1312.6199}.

\bibitem{Szegedy2014GoingDW}
Ch. Szegedy, W. Liu, Y. Jia, P. Sermanet, S. E. Reed, D. Anguelov, D. Erhan, V. Vanhoucke and A. Rabinovich (2014) Going deeper with convolutions,\textit{IEEE Conference on Computer Vision and Pattern Recognition (CVPR 15)}, p. 1-9.

\bibitem{TarasenkoWaves2011}
S. S. Tarasenko (2011) A General Framework for Development of the Cortex-like Visual Object Recognition System: Waves of Spikes, Predictive Coding and Universal Dictionary of Features, \textit{Proceedings of The 2011 International Joint Conference on Neural Networks}, p. 1515-1522.

\bibitem{Thorpe1996SpeedOP}
S. I. Thorpe and D. Fize and C. Marlot (1996) Speed of processing in the human visual system, \textit{Nature}, 381, p. 520-522.

\bibitem{Thorpe2001SpikebasedSF}
S. J. Thorpe, A. Delorme and R. van Rullen (2001) Spike-based strategies for rapid processing, \textit{Neural networks : the official journal of the International Neural Network Society}, 14 (6-7), p. 715-725.

\bibitem{VanRullen2002SurfingAS}
R. van Rullen and S. J. Thorpe (2002) Surfing a spike wave down the ventral stream,\textit{Vision Research}, 42, p. 2593-2615

\bibitem{Varghese2018ChangeNetAD}
A. Varghese and J. Gubbi and A. Ramaswamy and P. Balamuralidhar (2018) ChangeNet: A Deep Learning Architecture for Visual Change Detection, \textit{ECCV 2018 Workshops}. 

\bibitem{Zagoruyko2015LearningTC}
S. Zagoruyko and N. Komodakis (2015) Learning to compare image patches via convolutional neural networks, \textit{2015 IEEE Conference on Computer Vision and Pattern Recognition (CVPR 15)}, p. 4353-4361.

\end{thebibliography}

}

\end{document}